\documentclass{article} 
\pdfoutput=1
\usepackage[utf8]{inputenc}

\usepackage{iclr2018_workshop,times}
\usepackage{url}
\usepackage{graphicx}
\usepackage{multirow}
\usepackage{lipsum}
\usepackage{amsthm}
\usepackage{mathtools}
\usepackage{xfrac}
\usepackage[export]{adjustbox}
\usepackage{longtable}
\usepackage{booktabs}
\usepackage{colortbl}
\usepackage{xcolor}

\usepackage{hyperref}
\hypersetup{
    colorlinks=true,
    urlcolor=[rgb]{0.188, 0.498, 0.886},
    linkcolor=[rgb]{0.188, 0.498, 0.886},    
    filecolor=magenta,      
    citecolor=[rgb]{0.188, 0.498, 0.886},
}

\usepackage{algorithm} 
\usepackage{program} 
\usepackage{algorithmic}
\usepackage{listings}
\usepackage{geometry}

\usepackage{caption,setspace}
\DeclareCaptionFormat{listing}{\rule{\dimexpr0.9\columnwidth+17pt\relax}{0.4pt}\par\vskip{1pt}#1#2#3}
\newcounter{code}

\makeatletter
\newenvironment{code}[1][htb]{%
    \renewcommand{\ALG@name}{Exemplary vulnerability} 

   \begin{algorithm}[#1]%
  }{\end{algorithm}}
\makeatother

\DeclareMathVersion{sans}
\SetSymbolFont{operators}{sans}{OT1}{cmbr}{m}{n}
\SetSymbolFont{letters}  {sans}{OML}{cmbrm}{m}{it}
\SetSymbolFont{symbols}  {sans}{OMS}{cmbrs}{m}{n}

\lstnewenvironment{sflisting}[1][]
  {\lstset{#1}\mathversion{sans}}{}

\usepackage[normalem]{ulem}

\definecolor{grayalias}{HTML}{3F4444}

\definecolor{bluealias}{HTML}{307FE2}

\title{Towards an open standard for assessing the severity of robot security vulnerabilities, the Robot Vulnerability Scoring System (RVSS).}

\author{Víctor Mayoral Vilches,
  \textbf{Endika Gil-Uriarte},\\
  \textbf{Irati Zamalloa Ugarte},
  \textbf{Gorka Olalde Mendia},\\
  \textbf{Rodrigo Izquierdo Pisón},
  and \textbf{Alejandro Hernández Cordero}\\
  Alias Robotics S.L. \\
  Vitoria-Gasteiz, Álava, Spain \\
  \texttt{victor@aliasrobotics.com} \\
  \And
  \textbf{Lucas Apa} and \textbf{César Cerrudo} \\
  IOActive Inc. \\
  Seattle, Washington, USA
}

\newcommand{\ra}[1]{\renewcommand{\arraystretch}{#1}}

\usepackage{pdfpages}

\begin{document}

\maketitle

\vspace{-1em}
\begin{abstract}
Robots are typically not created with security as a main concern. Contrasting to typical IT systems, cyberphysical systems rely on security to handle safety aspects. In light of the former, classic  scoring methods such as the Common Vulnerability Scoring System (CVSS) are not able to accurately capture the severity of robot vulnerabilities. The present research work focuses upon creating an open and free to access Robot Vulnerability Scoring System (RVSS) that considers major relevant issues in robotics including a) robot safety aspects, b) assessment of downstream implications of a given vulnerability, c) library and third-party scoring assessments and d) environmental variables, such as time since vulnerability disclosure or exposure on the web. Finally, an experimental evaluation of RVSS with contrast to CVSS is provided and discussed with focus on the robotics security landscape.

\end{abstract}



\section{Introduction}
\label{sec:intro}

We are witnessing the rapid dawn of the robotics industry, but robots are seldom created with security as a concern. There is an emerging need to define formal security assessment methods for robotics, addressed in some of our previous research \cite{RSF} due to the vast attack surface they expose. Some pioneering work has focused onto generating primary concerns and social awareness on robot security faults \cite{hackingbeforeskynet, hackingbeforeskynet2}, including those generated by cyber attackers. From consumer and entertainment robots used in homes, to those used on assembly lines working closely with humans (collaborative robots or \emph{cobots}) to those used in healthcare, robot security is to be set as an emerging priority  \cite{2018arXiv180606681A}.

Nevertheless, security in robotics is often mistaken with safety. The integration between these two areas from a risk assessment perspective was studied in \cite{1673343, 2018arXiv180606681A} which resulted in an unified security and safety risk framework.
Commonly, robotics safety is understood as developing protective mechanisms against accidents or malfunctions, whilst security is aimed to protect systems against risks posed by malicious actors \cite{safetysecurity}. A slightly alternative view is the one that considers safety as protecting the environment from a given robot, whereas security is about protecting the robot from a given environment.

Previous work \cite{RSF} evidences that a wide range of robot-attacks are conceivable, well beyond network attacks. Such attacks require identifying of weaknesses in systems, which are embodied into concrete \emph{known} or unknown i.e. \emph{Zero-day} vulnerabilities for a given robot, which may be exploited by an attacker. However, our research has found that classification of robot vulnerabilities seldom follows commonly held standards. 

Most frequently, rating of vulnerabilities is achieved by means of the Common Vulnerability Scoring System (CVSS) \cite{mell2006common}, an accessible open system. Currently on version 3.0, CVSS is subject to continuous revision and improvement as an  effort to adapt to the evolution of technology. Nevertheless, estimating the risk to non-traditional computing devices was not on the original conception of CVSS, therefore, its application to robot vulnerabilities is inadequate.

Inspired by the state of the art described above, the present paper is aimed to shed some light into robot vulnerabilities and appropriate classification of their severity. The goal of our work is to present, evaluate and stimulate discussion around our proposition: a specific rating system for robot vulnerabilities able to capture the crucial features of robot vulnerabilities and able to produce numerical scores reflecting accurately its severity. Herein, we release version1 of the Robot Vulnerability Scoring System (RVSS). 

Our research is founded upon previous scoring systems \cite{mell2006common}, reported robot security requirements \cite{RSF} and the few and laudable well identified and reported robot threats and vulnerabilities \cite{hackingbeforeskynet, hackingbeforeskynet2}. Following well adopted and standardized approaches, we propose a vector creation scheme analog to CVSS, able to provide a numerical score that could then be translated into a qualitative representation. Furthermore, we provide an experimental evaluation of robot vulnerabilities and discuss the contrasting viewpoints between the traditional CVSS and the presented RVSS vulnerability scoring systems. The final goal of our work is to help robotics organizations to reliably assess and prioritize their vulnerability management processes, enabling appropriate mitigation.

The remainder of the paper is structured as follows: Section \ref{sec:previous} describes prior art in the field of (robot) vulnerability discovery and cataloguing. Section \ref{sec:rvss} elaborates upon RVSS. Section \ref{sec:CVSSvsRVSS} provides an experimental evaluation framework for comparison of RVSS to the IT-based CVSSv3 and discuss the outputs. The final epigraph \ref{sec:conclusions}, elaborates upon  major conclusions drawn from the present work.

\section{Previous work}
\label{sec:previous}
Even if there is an increasing concern about robot vulnerabilities on social media, news and academic literature, very little comprehensive research efforts have been conducted into logging and cataloging them. This lack may well be because a reference security assessment scheme was missing, to date. Our previous work highlights the needs for systematic security assessment in robotics, and provides with a reference procedure \cite{RSF}. Nonetheless, further research evidenced that the few reported vulnerabilities in robots are difficult to catalogue and severity assessment is substantially different from conventional vulnerabilities. Fitting and justifying accurately a given robot vulnerability into conventional scoring systems, such as the Common Vulnerability Scoring System (CVSS), is a complex task.

In short, CVSS provides a method to capture the main characteristics of a vulnerability and produce a numerical score reflecting its severity in a given context. The numerical score can then be transposed into a qualitative representation (such as low, medium, high and critical) that builds vector representations providing an overall score, indicating severity. These risk scores are used to help organizations to properly address, evaluate and prioritize the security of their products or assets.


CVSS was created by the National Infrastructure Advisory Council (NIAC) in the early 2000s, effort that led to the publication of CVSS version 1 (CVSSv1) in 2005, with the aim to "provide open and universally standard severity ratings of software vulnerabilities" \cite{cvssv1}. Currently its use is mandated or recommended for assessments in different security-critical domains, from medical devices to the credit card industry \cite{RUOHONEN2017}, and its fundamentals are continuously reviewed. The first CVSS draft was not subject any sort of peer review or review by other organizations but the creator. For the years to follow, NIAC appointed the Forum of Incident Response and Security Teams (FIRST) to further develop the CVSS. Thereafter, feedback from stakeholders utilizing CVSSv1 in production arose "significant issues with the initial draft of CVSS" which motivated work on CVSS version 2 (CVSSv2), released in 2007 \cite{cvss2}. 

CVSSv2 is composed of three metric groups: Base, Temporal, and Environmental; each consisting of a set of metrics. Further user feedback and stakeholder iteration motivated version 3 in 2012, ending with CVSSv3.0 being released in June 2015; which maintained the v2 backbone, but added some changes with regard to downstream effects of vulnerabilities. Newly proposed changes included novel metrics such as Scope (S) and User Interaction (UI), other former metrics like Authentication were changed or parsed to newer notation such as Privileges Required (PR).

One of the main changes to v3 was to reconsider the meaning of “Scope” as the feasibility of a single vulnerability impacting resources beyond its privileges \cite{cvss2}, with particular focus on IT environments. We find it extremely relevant that, although CVSS is a public standard \cite{cvssv1} adopted by security on a worldwide basis, it was never meant to address the increasing trend in connected devices, such as the ones connected to the IoT. Some works, such as \cite{cvssiot} have previously indicated that CVSS frameworks are not adequate for cyberphysical system vulnerability scoring, and highlighted the need of further refining accurate Environmental and Temporal metrics that provide the needed amount of detail about each vulnerability use case. Similarly, it is not the goal of the CVSS scheme to address any kind of physical influence onto the environment, such as \emph{Safety}. Needless to say, robots closely interact with the environment and human beings, and well deserve a different consideration to beyond conventional technology. Nonetheless, CVSS is showing a degree of use for IoT devices and some cyberphysical systems. For the particular case of robotics, only a few pioneering occasions have further catalogued disclosed vulnerabilities in Common Vulnerability Exposure (CVE archive), such as the two existing vulnerability reports related to the VgoRobot (Vecna). To date, this is one of the limited exemplary occasions where the CVEs are reserved for a robot related vulnerability. 

In our opinion, the current lack of use of the CVE scheme is motivated by two factors. The first one is the reluctance of most robot manufacturers to display signs of weakness in public, at the dawn of the industry with still poor normalization and standardization. Although robot vulnerabilities can easily be attributed within a Common Weakness Enumeration category or ID, no further CVE has been reserved by stakeholders. The second but not less important one is the fact that CVE and CVSS are intimately ligated. Since the use of CVSS is highly arguable for robot vulnerabilities, it does not provide additional value to continue logging vulnerabilities within a conceptually incomplete/inappropriate scheme. It is conceivable that further research will be devoted to normalization of outputs of RVSS, particularly related to the constitution of a Robot Vulnerability Exposure archive.



Additionally, we found that most of the common-IT based vulnerabilities reported within the CVE scheme seldom report further Temporal and Environmental metrics, which provide poor use-case contextualization of the given vulnerabilities. This lack yields a rather basic and abstract assessment of vulnerabilities that does not capture the full relevance of the detected vulnerability. We find it crucial to accurately produce the former for a given robot deployment (use-case) scenario. 


According to CVSSv3.0, vulnerability vectors are constructed as follows:


\begin{equation}
  \label{eq:1}
  CVSS:3.0/AV:N/AC:L/PR:N/UI:N/S:U/C:H/I:N/A:N
\end{equation}

A vulnerability that may score 7.5 our of 10 in the CVSSv3.0 may still be relatively low rated, even if the outcomes of the exploited vulnerabilities can led to the most explicit adverse \emph{Safety} outcomes conceivable.

Hence, after thorough examination of CVSS, we conclude that this existing system is not suitable for the peculiarities of robotics landscape. \emph{Safety} implications should be addressed necessarily. Thereafter, downstream vulnerability implications (e.g. privilege escalation etc) are to be addressed, with particular regard to the aforementioned security concerns. Indications of exploitability are to be developed further. For example, the period of time since vulnerability was indicated (Age metric) needs to be carefully observed.
Other relevant consequences/drivers of democratization of technology should be addressed for robotics, such as the use of third party libraries, collaborative software development efforts, and sharing of hacks. For example, aspects such as Web and Deep Web exposure and Social Networks indicators are often underestimated, however, reliable quantification of such variables may be too ambitious for the purpose of the present research. Similarly, Environmental features such as Environment specific outcomes are to be deeply reviewed. Our inputs are discussed deeply in Section \ref{sec:rvss}.


\section{The Robot Vulnerability Scoring System}
\label{sec:rvss}

\subsection{Reviewing CVSS version 3 for robotics}

Inspired by the previous work highlighted in Section \ref{sec:previous}, we have re-elaborated on top of the third version of the Common Vulnerability Scoring System (CVSS). The system is based upon three main \emph{metric groups} that should be considered in robot-related vulnerabilities. Each of them is subsequently divided in particular \emph{subgroups}, which organize together specific \emph{metric names} to be taken into account, some of which are mandatory, whilst other remain optional. \emph{Possibles values} are assigned to each metric related attributions. The CVSSv3 structure is summarized in Table \ref{table:cvss3} within Appendix \ref{appendix:cvss3summary}. The mathematical foundations of CVSSv3 are captured in Appendix \ref{appendix:cvss3math}.




    The Listing below provides a vulnerability example based on CWE-862 that applies to Robotis\footnote{\url{http://www.robotis.com/}} robots running Firmware OP2 with versions from 2015-03-26.

\begin{code}
\caption{Connection via established control port.}
According to \cite{hackingbeforeskynet2}, missing authorization mechanisms in Robotis RoboPlus protocol allow remote attackers to gain unauthorized control of the robots via network communication. This vulnerability applies to many of the robots running the affected firmware and could be captured with the CVSS vector expressed in Equation \ref{eq:1}:

\begin{equation}
  \label{eq:1}
  CVSS:3.0/AV:N/AC:L/PR:N/UI:N/S:U/C:N/I:H/A:H
\end{equation}

\end{code}



Following  \cite{mell2006common}, Equation \ref{eq:1} can be better understood as:

\begin{equation}
  \label{eq:2}
  \begin{split}
    CVSS:3.0/\underbrace{AV:N}_\text{Attack Vector (AV) = N = 0.85}/\underbrace{AC:L}_\text{Attack Complexity (AC) = L = 0.77} \\
    /\underbrace{PR:N}_\text{Privileges Required (PR) = N = 0.85}/\underbrace{UI:N}_\text{User Interaction (UI) = N = 0.85}/\underbrace{S:U}_\text{Scope (S) = U = Unchanged}\\
    /\underbrace{C:N}_\text{Confidentiality (C) = N = 0}/\underbrace{I:H}_\text{Integrity (I) = H = 0.56}/\underbrace{A:H}_\text{Availability (A) = H = 0.56}
  \end{split}
\end{equation}

Putting it all together with the mathematical expressions to derive CVSS version 3 scores described in Appendix \ref{appendix:cvss3math}:

\begin{equation}
  \label{eq:3}
  \begin{split}
    CVSS:3.0/\underbrace{\underbrace{\underbrace{AV:N}_\text{0.85}/\underbrace{AC:L}_\text{0.77}/\underbrace{PR:N}_\text{0.85}/\underbrace{UI:N}_\text{0.85}}_\text{Exploitability = 8.22$\cdot$0.85$\cdot$0.77$\cdot$0.85$\cdot$0.85 = 3.8870}
    /\underbrace{\underbrace{S:U}_\text{Unchanged}/\underbrace{\underbrace{C:N}_\text{0}/\underbrace{I:H}_\text{0.56}/\underbrace{A:H}_\text{0.56}}_\text{1-[(1-0)$\cdot$(1-0.56)$\cdot$(1-0.56)]=0.80640}}_\text{Impact = 6.42$\cdot$0.80640 = 5.1771}}_\text{$Base_{score}$  = roundup(min[(3.8870 + 5.1771), 10]) = roundup(9.0641) = 9.1}
  \end{split}
\end{equation}

Equations \ref{eq:1}, \ref{eq:2} and \ref{eq:3} provide a mathematical walkthrough of the vulnerability vector deriving into a score of \textbf{9.1} which according to the qualitative severity scoring rate shown in Table \ref{table:cvss3severity} corresponds with the \emph{critical range}.

This scoring may be modified by introducing the \emph{Temporal} subgroup metrics. Equation \ref{eq:eq4} illustrates the addition of this metric subgroup. After adding the corresponding values, the scoring of the same vulnerability lowers to \textbf{8.6}, lowering severity score from \emph{Critical} to \emph{High} according to Table \ref{table:cvss3severity} severity scoring rates.

\begin{equation}
  \label{eq:eq4}
    \resizebox{.9\hsize}{!}{$
    CVSS:3.0/\underbrace{\underbrace{\underbrace{\underbrace{AV:N}_\text{0.85}/\underbrace{AC:L}_\text{0.77}/\underbrace{PR:N}_\text{0.85}/\underbrace{UI:N}_\text{0.85}}_\text{Exploitability = 8.22$\cdot$0.85$\cdot$0.77$\cdot$0.85$\cdot$0.85 = 3.8870}
    /\underbrace{\underbrace{S:U}_\text{Unchanged}/\underbrace{\underbrace{C:N}_\text{0}/\underbrace{I:H}_\text{0.56}/\underbrace{A:H}_\text{0.56}}_\text{1-[(1-0)$\cdot$(1-0.56)$\cdot$(1-0.56)]=0.80640}}_\text{Impact = 6.42$\cdot$0.80640 = 5.1771}}_\text{$Base_{score}$  = roundup(min[(3.8870 + 5.1771), 10]) = roundup(9.0641) = 9.1}/\underbrace{\underbrace{E:P}_\text{0.94}/\underbrace{RL:U}_\text{1.0}/\underbrace{RC:C}_\text{1.0}}_\text{Temporal}}_\text{Temporal = roundup(9.1$\cdot$0.94$\cdot$1.0$\cdot$1.0) = roundup(8.5540) = 8.6}
    $}
\end{equation}

CVSS version 3 allows to further contextualize the score by introducing the \emph{Environment} group metrics as a way to reflect the importance of the affected asset to the user organization. These metrics enable the analyst to customize score depending on the importance of different aspects such as confidentiality, integrity or availability. Given the vulnerability described above, an analyst could make use of the \emph{Environmental} metrics by adding special factors: availability and integrity to prioritize these two aspects. Taking this into account, Equation \ref{eq:eq5} describes the derivation process of the new score which becomes \textbf{9.3}.

\begin{equation}
  \label{eq:eq5}
    \resizebox{.9\hsize}{!}{$
    CVSS:3.0/\underbrace{\underbrace{\underbrace{\underbrace{AV:N}_\text{0.85}/\underbrace{AC:L}_\text{0.77}/\underbrace{PR:N}_\text{0.85}/\underbrace{UI:N}_\text{0.85}}_\text{Exploitability = 8.22$\cdot$0.85$\cdot$0.77$\cdot$0.85$\cdot$0.85 = 3.8870}
    /\underbrace{\underbrace{S:U}_\text{Unchanged}/\underbrace{\underbrace{C:N}_\text{0}/\underbrace{I:H}_\text{0.56}/\underbrace{A:H}_\text{0.56}}_\text{1-[(1-0)$\cdot$(1-0.56)$\cdot$(1-0.56)]=0.80640}}_\text{Impact = 6.42$\cdot$0.80640 = 5.1771}}_\text{$Base_{score}$  = roundup(min[(3.8870 + 5.1771), 10]) = roundup(9.0641) = 9.1}/\underbrace{\underbrace{E:P}_\text{0.94}/\underbrace{RL:U}_\text{1.0}/\underbrace{RC:C}_\text{1.0}}_\text{Temporal}/\underbrace{\underbrace{IR:H}_\text{1.5}/\underbrace{AR:H}_\text{1.5}}_\text{Environmental}}_\text{Environmental = 9.3}
    $}
\end{equation}

The resulting score indeed reflects the intent of the analyst to increase the relevance of availability and confidentiality for this given vulnerability. However, we observe that the possibilities for the analyst to modify the score and adapt it to the real environment are rather limited. Particularly when dealing with cyberphysical systems, such as robots.

After thorough exploration of the possibilities that CVSS version 3 offers for a particular example in robotics, the following questions arise: \emph{Provided that this vulnerability applies to a robot, can the analyst modulate the scoring based on the physical damages that the robot can potentially create to the environment (safety aspects)? Could the analyst include within the score collateral damages that the robot may cause to itself due to the exploitation of such vulnerability?  Can the age of a reported vulnerability influence its severity? Can the security analyst distinguish between attacks made through different networking channels or networks within the robot (internal network, external network, etc.)? Can the security analyst reflect the social impact of a given hazard caused by the vulnerability?} 

We find that the traditional metrics do not address these issues at all, and detect conflicts with the following metrics:

\begin{itemize}
    \item \emph{Attack Vector (AV)} measures the context by which vulnerability exploitation is possible (the larger the more remote) and can adopt the following values:
    \begin{itemize}
        \item \texttt{Network (N) = 0.85} 
        \item \texttt{Adjacent Network (A) = 0.62}
        \item \texttt{Local (L) = 0.55}
        \item \texttt{Physical (P) = 0.2}
    \end{itemize}
    This metric originally aimed to capture whether software products were vulnerable from \underline{a)} anywhere on the Internet (\texttt{Network}), \underline{b)} closeby (certain proximity), adjacent networks  (\texttt{Adjacent Network}) such as VPNs, \underline{c)} a local networking structure within the organization (\texttt{Local}) or \underline{d)} in physical contact with the machine where the product is running (\texttt{Physical}). 
    
    The weights were designed accordingly and assume that physical access is typically restricted (therefore weighted less) since software products are running in restricted, monitored server environments. This assumption, however, does not apply to all robots. While some may be working in industrial facilities with restricted physical access; some others may operate in public areas, where anyone can physically access the robots (e.g.: robot cars, infotainment robots, healthcare robots, etc.). Similarly, robots are typically designed with several -more than one, at least- networking layers. Typically, an external network -used for operating and monitoring the robot- and an internal network -where all distributed components and modules exchange information to compute the desired behavior-, are often accessible with physical contact.
    

    \item \emph{Scope (S)} metric measures the ability for a vulnerability in a software component to impact resources beyond its means or privileges. This becomes critical when dealing with robots and the adverse physical outcomes they may cause. In our opinion, this aspect is currently not properly contemplated. Equations \ref{eq:vuln2:1} and \ref{eq:vuln2:2} present an scenario where this aspect is highlighted.

    \item Through the \emph{Environmental} group of metrics and in particular, the modified versions of \emph{Confidentiality (C), Integrity (I), and Availability (A)} are provided. The analyst can adapt the impact of a successfully exploited vulnerability when the \emph{Scope (S)} has changed and escalated. These metrics empower analysts to fine tune the scoring with the options described in Table \ref{table:cvss3} within Appendix \ref{appendix:cvss3summary}. However, as recent research points out \cite{cvss2}, it is not a conventional analyst task to weigh and prioritize safety by assessing confidentiality, integrity and availability. We agree in acknowledging that the potential hazards cannot be described through these metrics, but at least, reformulation is needed for practical purposes.

\begin{code}
\caption{OBD II dongle vulnerability, modern cars hack.}
\cite{cvss2} describes a vulnerability in an OBD II dongle, which can connect to any modern car's (considered as robots due to their architecture) internal control software and operate them at desire. According to \cite{cvss2}: \emph{The intended use (of the OBD II dongle) is for your cell phone to communicate with the dongle over WiFi, and the dongle can use the phone's Internet connection if necessary. The device relays commands it receives over WiFi to the internal networks in the vehicle and vice-versa. It uses a weak and unsophisticated password: “password” to connect by default, which  can be changed through the devices mobile application, however, it is not mandatory.} \\

According to the authors, the result of exploiting this vulnerability could lead to an attacker potentially kidnapping the car, robbing the contents of the car, stealing data stored on it (i.e., contacts or GPS history), eavesdroping on the driver, or even intentionally causing the car to crash. The scoring obtained when evaluating this vulnerability with CVSSv3 is \textbf{7.6} (Equation \ref{eq:vuln2:2}) but whenever the \emph{Scope} is set to "changed"\footnote{due to the fact that the vulnerability applies to the overall robot (or vehicle)}, the scoring is increased to \textbf{8.8}. A mild variation of 1.2 points is shown in this particular scenario where the consequences could lead to human casualties, indicating that the scoring system is clearly failing to capture the security and safety critical aspects of this example.

\begin{equation}
  \label{eq:vuln2:1}
  CVSS:3.0/AV:A/AC:L/PR:N/UI:N/S:C/C:L/I:L/A:H = 8.8
\end{equation}

\begin{equation}
  \label{eq:vuln2:2}
CVSS:3.0/AV:A/AC:L/PR:N/UI:N/S:U/C:L/I:L/A:H = 7.6
\end{equation}
\end{code}    
    
\end{itemize}

\subsection{Proposed improvements for robotics}

With the aim of stimulating a discussion and inspired by previous work \cite{cvssiot, cvss2}, we propose a set of suggestions to re-design the metrics and numerical attributes of CVSS. We have followed a conservative approach when reviewing the metrics, however, as highlighted by several authors \cite{cvssiot, cvss2, vulnstudy}, we find it extremely relevant to cooperate in the formulation of a consistent scoring system for the robotics domain. Overall, we propose to the community of robotics security researchers the following changes:



\begin{itemize}
    \item \textbf{Base metric} group should be modified in the following terms:
    \begin{itemize}
        \item Modify the \emph{Attack Vector (AV)} to add metrics that are \emph{able to distinguish attacks available through different network types within a robot} and include \emph{different levels of physical access}. 
        
        \item Introduce a new metric named Age (Y) that averages the relevance of a vulnerability given the time span since vulnerability was first reported. This metric further allows the consideration of the time-frame initiated from the point where a vulnerability is disclosed.
        


        \item Include a \emph{Safety} metric on the \emph{Impact} subgroup that addresses the safety-related hazards of the vulnerability. The metric should be able to measure potential physical outcomes onto human or environmental safety, once the vulnerability is exploited, which requires a separate consideration for each particular robot.

        \item As reported by some security researchers \cite{vulnstudy}, \emph{"response in social media gives an indication of how important the vulnerability is as opposed to the score represented by the CVSS"}. We recognize it is an extremely important component of vulnerabilities, which captures impacts of cybersecurity democratization. Even if we will not take any actions on this regard due to the substantial complexity of the assessment, we reckon it may deserve further research. This does not preclude future additions of a \emph{Social Exposure} metric to the \emph{Impact} subgroup and provides a point for further discussions.
        
        %
        %
        %
    \end{itemize}



    \item Change \textbf{Environmental} metrics to the robotics-specific context as follows:
    \begin{itemize}
        \item Include \emph{Safety} as part of the modifiers available to analysts. This will allow analysts to re-evaluate context specific safety-related aspects.
        
        \item Modify the impact of changed \emph{Scope} in the mathematical expressions.
        
        \item Include guidance on cataloguing vulnerabilities in libraries (or similar). This is specially relevant in robotics, where middleware frameworks like ROS, YARP  and several other facilitate the task of putting together robot software, through collaborative Open Access platforms/libraries.

        
    \end{itemize}
\end{itemize}

The following subsection will elaborate into these proposed improvements by introducing a new scoring system to evaluate vulnerabilities for robot systems.

\subsection{An improved scoring architecture for robots: RVSS}

\begin{figure}[h!]
    \centering
    \includegraphics[width=\textwidth]{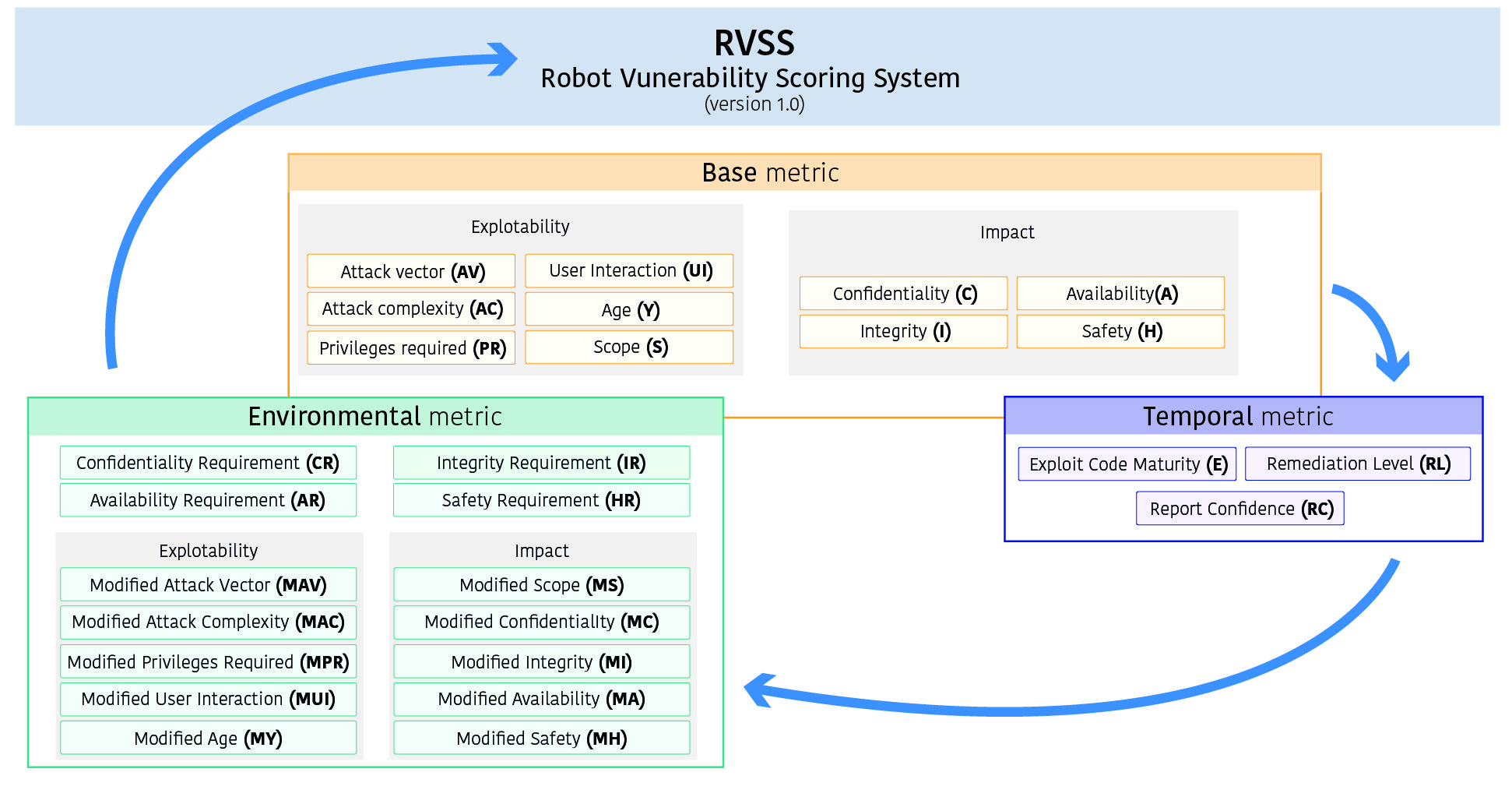}
    \caption{\footnotesize Components of Robot Vulnerability Scoring System version 1.0 (RVSS). Figure show metric groups (\emph{Base, Temporal} and \emph{Environmental} that are subsequently classified in subgroups. Each metric name is depicted in boxes within subgroups.}
    \label{fig:rvss}
\end{figure}

The present content is aimed at developing the Robot Vulnerability Scoring System (RVSS) as an extension of CVSSv3 for robotics. Figure \ref{fig:rvss} pictures the entire structure of the RVSS, a complete scoring system for robot vulnerabilities.

Below, we describe the changes proposed in detail for RVSS:

\begin{itemize}

    \item \textbf{Attack Vector (AV)} metric has been modified to be able to distinguish attacks coming through different network types within a robot. Mainly, we classify attack vectors in two subtypes: network and physical. Within the network types, we distinguish two subdivisions: external and internal robot network attack vectors: 
    \begin{itemize}
        \item \underline{Network Attack Vectors}:
        \begin{itemize}
            \item[o] \emph{External Robot Network Attack Vectors}:
            \begin{itemize}
                \item \texttt{Remote Network (RN) = 0.85}: attack vectors over a general area network such as the Internet.
                \item \texttt{Adjacent Network (AN) = 0.62}: vectors from networks within certain proximity, adjacent. The typical example is a corporate Virtual Private Network (VPN).
            \end{itemize}
            \item[o] \emph{Internal Robot Network Attack Vectors}:
            \begin{itemize}
                \item \texttt{Internal Network (IN) = 0.4}: vectors of attack that come from within the internal network of the robot. These internal networks typically contain sensors, actuators and other robot modules and components. Vectors targeting these networks can be catastrophic however, as most robots do not expose their internal networks, these vectors of attack will typically require some sort of physical access to the internal network.             
            \end{itemize}
        \end{itemize}
        
        \item \texttt{\underline{Local} (L) = 0.55}: A vulnerability exploitable with local access means that the vulnerable robot or robot component is not bound to the network stack, and the attacker’s path is via read/write/execute capabilities. In some cases, the attacker may be logged-in locally in order to exploit the vulnerability or may rely on User Interfaces to execute a malicious file.
        \item \underline{Physical Attack Vectors}:
        \begin{itemize}
            \item[o] \texttt{Physical Public (PP) = 0.62}: attack vectors coming from physical contact with a robot that is deployed in publicly accessible areas. An information and entertainment (infotainment) robot deployed in an airport or a robot car parked in a public area are two of the possible examples.
            
            \item[o] \texttt{Physical Restricted (PR) = 0.4}: attack vectors coming from physical contact with a robot that is deployed in areas restricted to the general public, yet accessible to certain individuals within an organization or subgroup. Representative examples include surgical robots exposed to the personnel within a hospital, warehouse automation robots operating within the scope of a given corporation's facility or a collaborative robot automating a particular task within an industrial work-space  with restricted access. 
        
            \item[o] \texttt{Physical Isolated (PI) = 0.2}: vectors of attack coming from physical contact with robots that are operating in an isolated manner and without human intervention or proximity. These robots operate in an isolated manner and only selected individuals have the clearance to access them physically. Among the possible scenarios, a representative example could be a robot operating in an industrial automotive facility executing a sub-task within the complete assembly of a car. Such robot operates in an isolated manner and as part of a manufacturing chain.
            
        \end{itemize}
    \end{itemize}
    In addition to this novel structure and in order to increase the descriptive capability of this metric, we propose the inclusion of more than one \emph{possible values} within the \emph{Attack Vector (AV)} metric. That is, for cases such as an attack vector executed over the internal network of a robot that in addition, requires physical contact. Provided such robot is available in public spaces, we envision that the attack vector metric should not be calculated as: $AV = IN = 0.2$ since this would be representative of an attack vector simply performed over an internal network through whatever means the attacker may unfurl. Instead, we could increase the resolution of the attack vector metric by using the product of both conditions delivering $AV = IN \cdot PP = 0.2 \cdot 0.62 = 0.124$. The resulting value is lower than the one obtained from considering simply an internal network attack. To us, the procedure reflects the additional complexity required by certain robot attack vectors, which demand more than one action or step for the attacker.\\
    
    The new proposed syntax for this metric is presented in Equation \ref{eq:syntaxav} allowing compositions like $AV:ANPI$:
    
\begin{equation}
    \label{eq:syntaxav}
    \resizebox{.6\hsize}{!}{$
    AV = PossibleValue_1 \bigg[PossibleValue_2 [ PossibleValue_N ]\bigg]
    $}
\end{equation}    
    
    The formula used to calculate the new proposed metric for the Attack Vector is described in Equation \ref{eq:mathav}. In the previous case, $AV:ANPI$, the new proposed value for this metric would result in $AV=0.62 \cdot 0.2 = 0.124$.
    
\begin{equation}
    \label{eq:mathav}
    \resizebox{.8\hsize}{!}{$
    AV = PossibleValue_1 \text{ (e.g. IN=0.2) } \bigg[\cdot PossibleValue_2 \text{ (e.g. PP=0.62) } [\cdot PossibleValue_N ]\bigg]
    $}
\end{equation}    

    our reference implementation\footnote{\url{https://github.com/aliasrobotics/RVSS}} provides only a limited amount of combinations, extensions will be proposed in the future as they are required.

    \item \textbf{Age (Y)}: is a new metric that captures the time span since vulnerability was first reported (in years). It applies particularly to vulnerabilities not previously reported within vulnerability scoring schemes. The possible values are listed below:

    \begin{itemize}
        \item \texttt{Zero Day (Z) = 1.0}: Recent vulnerabilities with a time frame of a maximum of one month since it was first reported.
        \item \texttt{1 year or less (O) = 1.1}: One year or less since the vulnerability was first reported.
        \item \texttt{Less than 3 years (T) = 1.2}: Between one and three years since the vulnerability was first reported.
        \item \texttt{More than 3 years (M) = 1.5}: Old vulnerabilities considered with an age (in years) of more than 3 years since it was reported.
        \item \texttt{Unknown (U) = 1.0}: In those cases where the age is unknown we will use this multiplication factor.
    \end{itemize}
    
    It is a common practice that robot vulnerabilities are not adequately patched by robot stakeholders, but minor mitigation countermeasures are suggested/recommended to end users. This way, aim to we penalize older vulnerabilities which should, by the time being, have been diligently addressed by robot manufacturers. By introducing this new metric, the \emph{Explotaibility} metric subgroup is calculated as:

\begin{equation}
    Explotaibility = 8.22 \cdot AV \cdot AC
    \cdot PR \cdot UI \cdot Y
\label{eq:exploitability}
\end{equation}
    
    Accordingly, the RVSS vector is derived from Equation \ref{eq:agemod}:
    
\begin{equation}
  \label{eq:agemod}
  \resizebox{.8\hsize}{!}{$
  \begin{split}
    RVSS:1.0/\underbrace{AV:N}_\text{Attack Vector (AV)}/\underbrace{AC:L}_\text{Attack Complexity (AC)}
    /\underbrace{PR:N}_\text{Privileges Required (PR)} 
    /\underbrace{UI:N}_\text{User Interaction (UI)}
    /\underbrace{Y:U}_\text{\textbf{Age (Y)}}
    /\underbrace{S:U}_\text{Scope (S)} 
    /\underbrace{C:N}_\text{Confidentiality (C)}/\underbrace{I:H}_\text{Integrity (I)} 
    /\underbrace{A:H}_\text{Availability (A)} \Bigg[/\underbrace{E:U}_\text{Exploit Code Maturity (E)} \\
    /\underbrace{RL:U}_\text{Remediation Level (RL)}/\underbrace{RC:U}_\text{Report Confidence (RC)} \bigg[ /\underbrace{CR:X}_\text{Confidentiality Requirement (CR)}/\underbrace{IR:X}_\text{Integrity Requirement (IR)}/\underbrace{AR:X}_\text{Availability Requirement (AR)}/\underbrace{MAV:X}_\text{Modified Attack Vector (MAV)} 
    /\underbrace{MAC:X}_\text{Modified Attack Complexity (MAC)} \\
    /\underbrace{MPR:X}_\text{Modified Privileges Required (MPR)} 
    /\underbrace{MUI:X}_\text{Modified User Interaction (MUI)}
    /\underbrace{MC:X}_\text{Modified Confidentiality (MC)}/\underbrace{MI:X}_\text{Modified Integrity (MI)}/\underbrace{MA:X}_\text{Modified Availability (MA)} \bigg] \Bigg]
  \end{split}
  $}
\end{equation}       
    
    \item \textbf{Safety (H)}: has been added to the \emph{Impact} subgroup as part of the \emph{Base} group of metrics. This element measures potential physical hazards on humans or on the environment that a robot vulnerability can cause. The values that this metric can adopt are:
    \begin{itemize}
        \item \texttt{Unknown (U) = 0.0}: in the case where safety impact is unknown, this \emph{Possible Value} ensures that the equation is left untouched.
        \item \texttt{None (N) = 0.0}: when a vulnerability causes no safety implications, the overall score isn't affected. 
        \item \texttt{Environmental (E) = 0.15}: if there are safety consequences that may affect the environment in a direct manner, the overall score should increase.
        \item \texttt{Human (H) = 0.35}: if safety aspects could affect humans directly, the \emph{Safety} metric should reflect it considerably. This coefficient, together with its mathematical expression does so.
    \end{itemize}

    When including \emph{Safety} metric, the \emph{Impact} subgroup value is calculated through Equations \ref{eq:impact4} and \ref{eq:impact3}:
    
    \begin{equation}
        ISC_{Base} = 1 - [(1 - C) \cdot (1 - I ) \cdot (1 - A)] + 1.2 \cdot H
    \label{eq:impact4}
    \end{equation}
    
    \begin{equation}
    \resizebox{.8\hsize}{!}{$
    Impact =
    \begin{cases}
      6.42 \cdot ISC_{Base}, & \text{if}\ Scope \text{ Unchanged}\\
      7.52 \cdot [ISC_{Base} - 0.029] - 3.25 \cdot [ISC_{Base} - 0.02]^{15}, & \text{if}\ Scope \text{ Changed}\\
    \end{cases}
    $}
    \label{eq:impact3}
    \end{equation}    

    Summing up, including the \emph{Safety} metric as described above, the new vector is constructed as described at Equation \ref{eq:safety_vector} (default values have been assigned for the purpose of demonstration):

\begin{equation}
  \label{eq:safety_vector}
  \resizebox{.8\hsize}{!}{$
  \begin{split}
    RVSS:1.0/\underbrace{AV:N}_\text{Attack Vector (AV)}/\underbrace{AC:L}_\text{Attack Complexity (AC)}
    /\underbrace{PR:N}_\text{Privileges Required (PR)} 
    /\underbrace{UI:N}_\text{User Interaction (UI)}
    /\underbrace{Y:U}_\text{Age (Y)}
    /\underbrace{S:U}_\text{Scope (S)} 
    /\underbrace{C:N}_\text{Confidentiality (C)}/\underbrace{I:H}_\text{Integrity (I)} 
    /\underbrace{A:H}_\text{Availability (A)}/\underbrace{H:U}_\text{\textbf{Safety (H)}} \\ \Bigg[/\underbrace{E:U}_\text{Exploit Code Maturity (E)}/\underbrace{RL:U}_\text{Remediation Level (RL)}/\underbrace{RC:U}_\text{Report Confidence (RC)} \bigg[ /\underbrace{CR:X}_\text{Confidentiality Requirement (CR)}/\underbrace{IR:X}_\text{Integrity Requirement (IR)}/\underbrace{AR:X}_\text{Availability Requirement (AR)}/\underbrace{MAV:X}_\text{Modified Attack Vector (MAV)} 
    /\underbrace{MAC:X}_\text{Modified Attack Complexity (MAC)} \\
    /\underbrace{MPR:X}_\text{Modified Privileges Required (MPR)} 
    /\underbrace{MUI:X}_\text{Modified User Interaction (MUI)}
    /\underbrace{MC:X}_\text{Modified Confidentiality (MC)}/\underbrace{MI:X}_\text{Modified Integrity (MI)}/\underbrace{MA:X}_\text{Modified Availability (MA)} \bigg] \Bigg]
  \end{split}
  $}
\end{equation}

    \item \textbf{Modified Safety (MH)}: according to the official CVSSv3 specification, modified metrics enable the analyst to adjust the Base metrics according to modifications that exist within the analyst’s environment. That is, if an environment has changed in a way which would affect its Exploitability, Scope, or Impact, then the environment can reflect this via an appropriately-modified, score. Modified Safety (MH) proposes a metric to capture environmental changes that are related to safety aspects. Similar to Confidentiality, Integrity or Availability, we propose the same values as the corresponding Base metric as well as \texttt{Not Defined} value. This modifies the previous CVSSv3 equations as follows:
    
The modified impact $M.Impact$ new score is defined as:

\begin{equation}
\resizebox{.8\hsize}{!}{$
M.Impact =
\begin{cases}
  6.42 \cdot ISC_{Modified}, & \text{if Modified}\ Scope \text{ Unchanged}\\
  7.52 \cdot [ISC_{Modified} - 0.029]  & \text{if Modified}\ Scope \text{ Changed}\\
  - 3.25 \cdot [ISC_{Modified} - 0.02]^{15} & \\
\end{cases}
$}
\label{eq:mimpact}
\end{equation}

where

\begin{equation}
\resizebox{.8\hsize}{!}{$
ISC_{Modified}= min\bigg(\Big[1 - (1 - MC \cdot CR)
\cdot (1 - MI \cdot IR)
\cdot (1 - MA \cdot AR) \cdot (1 - MH \cdot HR)\Big],0.915 \bigg)
+ 1.2 \cdot MH \cdot HR
$}
\end{equation}

    With this modification, RVSS vector is represented as:
    
\begin{equation}
  \label{eq:modified_safety_vector}
  \resizebox{.8\hsize}{!}{$
  \begin{split}
    RVSS:1.0/\underbrace{AV:N}_\text{Attack Vector (AV)}/\underbrace{AC:L}_\text{Attack Complexity (AC)}
    /\underbrace{PR:N}_\text{Privileges Required (PR)} 
    /\underbrace{UI:N}_\text{User Interaction (UI)}
    /\underbrace{Y:U}_\text{Age (Y)}
    /\underbrace{S:U}_\text{Scope (S)} 
    /\underbrace{C:N}_\text{Confidentiality (C)}/\underbrace{I:H}_\text{Integrity (I)} 
    /\underbrace{A:H}_\text{Availability (A)}/\underbrace{H:U}_\text{Safety (H)} \\ \Bigg[/\underbrace{E:U}_\text{Exploit Code Maturity (E)}/\underbrace{RL:U}_\text{Remediation Level (RL)}/\underbrace{RC:U}_\text{Report Confidence (RC)} \bigg[ /\underbrace{CR:X}_\text{Confidentiality Requirement (CR)}/\underbrace{IR:X}_\text{Integrity Requirement (IR)}/\underbrace{AR:X}_\text{Availability Requirement (AR)}/\underbrace{HR:X}_\text{\textbf{Safety Requirement (HR)}}/\underbrace{MAV:X}_\text{Modified Attack Vector (AV)} 
    /\underbrace{MAC:X}_\text{Modified Attack Complexity (MAC)} \\
    /\underbrace{MPR:X}_\text{Modified Privileges Required (MPR)} 
    /\underbrace{MUI:X}_\text{Modified User Interaction (MUI)}
    /\underbrace{MC:X}_\text{Modified Confidentiality (MC)}/\underbrace{MI:X}_\text{Modified Integrity (MI)}/\underbrace{MA:X}_\text{Modified Availability (MA)}/\underbrace{MH:X}_\text{\textbf{Modified Safety (MH)}} \bigg] \Bigg]
  \end{split}
  $}
\end{equation}   

    the Safety Requirement (HR) values are reflected in Table \ref{table:rvss} and use similar values to those of Confidentiality, Availability or Integrity.

    \item \textbf{Modified Age (MY)}: Similar to Modified Safety, we provide with a modified version of the Age metric so that the analyst is able to adjust the Base metrics according to environmental changes. Adding this, the resulting vector for RVSS version 1.0 is presented in Equation \ref{eq:modified_age_vector}:
    
\begin{equation}
  \label{eq:modified_age_vector}
  \resizebox{.8\hsize}{!}{$
  \begin{split}
    RVSS:1.0/\underbrace{AV:N}_\text{Attack Vector (AV)}/\underbrace{AC:L}_\text{Attack Complexity (AC)}
    /\underbrace{PR:N}_\text{Privileges Required (PR)} 
    /\underbrace{UI:N}_\text{User Interaction (UI)}
    /\underbrace{Y:U}_\text{Age (Y)}
    /\underbrace{S:U}_\text{Scope (S)} 
    /\underbrace{C:N}_\text{Confidentiality (C)}/\underbrace{I:H}_\text{Integrity (I)} 
    /\underbrace{A:H}_\text{Availability (A)}/\underbrace{H:U}_\text{Safety (H)} \\ \Bigg[/\underbrace{E:U}_\text{Exploit Code Maturity (E)}/\underbrace{RL:U}_\text{Remediation Level (RL)}/\underbrace{RC:U}_\text{Report Confidence (RC)} \bigg[ /\underbrace{CR:X}_\text{Confidentiality Requirement (CR)}/\underbrace{IR:X}_\text{Integrity Requirement (IR)}/\underbrace{AR:X}_\text{Availability Requirement (AR)}/\underbrace{HR:X}_\text{Safety Requirement (HR)}/\underbrace{MAV:X}_\text{Modified Attack Vector (MAV)} 
    /\underbrace{MAC:X}_\text{Modified Attack Complexity (MAC)} \\
    /\underbrace{MPR:X}_\text{Modified Privileges Required (MPR)} 
    /\underbrace{MUI:X}_\text{Modified User Interaction (MUI)}
    /\underbrace{MY:X}_\text{\textbf{Modified Age (MY)}}
    /\underbrace{MC:X}_\text{Modified Confidentiality (MC)}/\underbrace{MI:X}_\text{Modified Integrity (MI)}/\underbrace{MA:X}_\text{Modified Availability (MA)}/\underbrace{MH:X}_\text{Modified Safety (MH)} \bigg] \Bigg]
  \end{split}
  $}
\end{equation}       
    
    
    \item \textbf{Scoring for libraries and middleware}: Currently, CVSSv3 scores vulnerabilities in libraries and other third-party abstractions with 0 due to the fact that there is no direct impact from the exploitation of these vulnerabilities alone. Conversely, third party libraries are particularly relevant in robotics where middleware frameworks like ROS, ROS 2.0, Orocos, YARP  and several others facilitate the task of putting together robot software. 
    
    There has been some discussions about this topic which are summarized to some extent by Carlage\cite{cvss3improvements}. Steaming from these previous thoughts, we have evaluated the following strategies:
    
    \begin{enumerate}
        \item \emph{Remove the restriction of scoring 0 when Impact is zero}: While simple implement, this path would lead to inconsistent comparisons between RVSS and CVSSv3.
        \item \emph{Complement the Impact subgroup of metrics with a new metric}: Such metric would be dedicated to raise the Impact subgroup metric above zero so that the existing restrictions don't apply and the overall score of RVSS is not zero for libraries or middleware.
        \item \emph{Score according to the reasonable worst case}: Suggested originally by Carlage \cite{cvss3improvements}, this would imply rating the Impact subgroup of metrics by guessing how third party code (libraries) would affect and select scores for the reasonable worst case.
    \end{enumerate}
    
    We adopt measures 2 and 3 by including the \emph{Safety (H)} metric in the evaluation of the \emph{Impact} and by considering the worst possible scenario for each one of the \emph{Impact} metrics, respectively.

    
\end{itemize}


\section{Experimental evaluation of the scoring systems for robots: (CVSSv3 vs RVSS)}
\label{sec:CVSSvsRVSS}

In this section, we propose a series of examples that compare our scoring system together with CVSSv3. Examples are summarized in Table \ref{table:comparervsscvss} and discussed below.

\ra{2}
\begin{longtable}{|l p{7cm} c c |}
  \caption{\footnotesize Comparing RVSSv1.0 and CVSSv3.0. The scoring last two rows present the (Base, Temporal, Environmental) scores of RVSSv1.0 and CVSSv3.0 respectively. Each corresponding vector is captured as a \emph{footnote}.}
  \label{table:comparervsscvss}\\

    \toprule
\textbf{\#} & \textbf{Vulnerability description} & \textbf{RVSSv1.0} & \textbf{CVSSv3.0} \\
    \midrule
    \endfirsthead 
    \toprule
\textbf{\#} &\textbf{Vulnerability description} & \textbf{RVSSv1.0} & \textbf{CVSSv3.0} \\
    \midrule
    \endhead 

  1 & Missing authorization mechanisms in Robotis RoboPlus protocol allow remote attackers to gain unauthorized control the robots via network communication. & (7.7, 7.7, 7.7)\footnote{RVSS:1.0/AV:ANPR/AC:L/PR:N/UI:N/Y:T/S:U/C:N/I:H/A:H/H:E} & (9.1, 9.1, 9.1)\footnote{CVSS:3.0/AV:N/AC:L/PR:N/UI:N/S:U/C:N/I:H/A:H} \\
  
  \rowcolor{black!5} 2 & An attacker on an adjacent network could perform command injection &  (10, 10, 10)\footnote{RVSS:1.0/AV:AN/AC:L/PR:N/UI:N/Y:O/S:U/C:H/I:H/A:H/H:E} & (8.8, 8.8, 8.8)\footnote{CVSS:3.0/AV:A/AC:L/PR:N/UI:N/S:U/C:H/I:H/A:H}\\
  
  3 & An stack-based buffer overflow in Universal Robots Modbus TCP service could allow remote attackers to execute arbitrary code and alter protected settings via specially crafted packets. & (10,10,10)\footnote{CVSS:3.0/AV:N/AC:L/PR:N/UI:N/S:C/C:H/I:H/A:H} & (10,10,10)\footnote{RVSS:1.0/AV:AN/AC:L/PR:N/UI:N/Y:T/S:C/C:H/I:H/A:H/H:H}\\
  
  \rowcolor{black!5} 4 & Exemplary vulnerability in ROS 2.0 communication middleware: Launching on arm64 with Fast-RTPS with fat archive from 2018-06-21 never quits. &  (5.9, 5.9, 5.9)\footnote{RVSS:1.0/AV:AN/AC:L/PR:N/UI:N/Y:O/S:U/C:N/I:N/A:N/H:H} & (0, 0, 0)\footnote{CVSS:3.0/AV:A/AC:L/PR:N/UI:N/S:U/C:H/I:N/A:H}\\
    
\end{longtable}

\subsection{Case study A: Missing authorization mechanisms in Robotis RoboPlus protocol}

The first example (number 1 in Table \ref{table:comparervsscvss}), discusses a vulnerability that affects all Robotis robots that run the OP2 Firmware, version from 2015-03-26. According to the authors of the vulnerability \cite{hackingbeforeskynet2}, RoboPlus Motion's server binary starts listening on TCP port 6501 and allows clients to control and edit motion components on the robot. The software does not perform any authentication that requires a  user identity to control robot joints and changing actuator data. The assigned CVSSv3 score is of 9.1\footnote{No modifier applied thereby all three, Base, Temporal and Environmental metrics are equal}. 

Using RVSS, we adapt the vector to include the above proposed changes and the additional mandatory metrics for this new scoring system, namely Age (Y) and Safety (H). Some authors claim that if the robot uses an extra wireless interface (something that not all robots include and that represent an add-on for many), the vulnerable port could be exposed on that interface as well. This is the rationale behind the CVSSv3 rating of \texttt{Network (N)} for the Attack Vector (AV) metric. We modify the Attack Vector (AV) to capture the fact a) these robots are not typically connected to remote networks but instead, to local (adjacent) networks,  b) physical access is typically needed to extend the robot and enable the wireless mechanisms and c) these robots are not publicly available but instead are available in restricted areas such as homes. We thereby use the rating \texttt{Adjacent Network and Physical Restricted (ANPR)} which delivers a preliminary Base score of 6.4 for RVSS, significantly lower than the CVSS one.

However, RVSS considers Age as metric for the Base group: The vulnerability was first reported on March 2017, thereby we rate the Age (Y) with \texttt{Less than 3 years (T) = 1.1} which elevates RVSS score to 6.6. 

The robots affected by the vulnerability are mostly toys or educational robots with limited physical action capability. Some could argue that these robots can hardly present any safety issue for the environment or for human beings. We adopt a conservative attitude and define that the safety aspects of these vulnerability can affect the environment, thereby the Safety (H) metric adopts a value of \texttt{0.15} (corresponding with Environmental (E) safety compromise) which influences the overall score resulting on 7.7.

The resulting RVSS score of 7.7 is significantly lower than the 9.1 from CVSSv3. While being conservative, it lowers the severity one complete level, from Critical to High. Moreover, discarding the safety implications we assumed, the RVSS score would lower another level, decreasing to Medium in the severity scale of Table \ref{table:cvss3severity}.

\subsection{Case Study B: Command injection in telepresence robot}

The second vulnerability of Table \ref{table:comparervsscvss} analyzes a flaw in Vecna's VGo Robot that applies to versions prior 3.0.3.52164. Vulnerable versions include: 3.0.3, 3.0.2, 2.1.0, 2.0.0, 1.5.5, 1.5.0, 1.4.2. According to the original report, an attacker on an adjacent network could perform command injections. As displayed in the table above, its CVSSv3 score is 8.8 which falls in the "High" severity section of CVSSv3.

We transform the vulnerability vector into RVSS. The sole addition of the Age (Y) metric, scored \texttt{Less than 1 year (O) = 1.1} results in a RVSS score of 9.0. The robot itself, a telepresence machine mounted on a mobile base with an apparent differential drive system presents little safety implications for either humans or the environment. These robots are typically employed in controlled public or private areas where they interact with humans. Their velocity is typically restricted by design which lowers the potential adverse safety outcomes.  A conservative evaluation would rate the Safety (H) metric with Environmental (E) implications resulting in a total score of 10.0, the maximum score. 

\subsection{Case Study C: Specially crafted Modbus network packet}

The third study case refers to Universal Robots	UR3, UR5, UR10. As described at \cite{hackingbeforeskynet2}, by exploiting a buffer overflow remotely in the Universal Robots Modbus TCP service and bypassing OS protections through a Return Oriented Programming (ROP) attack, an attacker is able to gain root access to the robot controller. From this point, the robot could be commanded as desired by commanded unauthenticated services, or even disable completely the safety mechanisms causing potential human harm. \emph{Note that the scope in the vector is set to \texttt{changed} since its implications go beyond the subsystem exploited itself.}

Both, CVSSv3 and our RVSS result in a score of 10.0.

\subsection{Case Study D: Vulnerabilities in robot middleware (libraries)}

As it is, the current CVSSv3 neglects the interpretation of vulnerabilities in code that provides functionality to other code, but which cannot run by itself. Libraries are an example. As such, Carlage\cite{cvss3improvements}  claims that libraries would not run without other code, resulting in a vulnerable library scoring value of 0 in  CVSSv3, asserted upon the fact that the attacker should not be able to access it. 

Let's analyze a practical example originally reported at \url{https://github.com/ros2/launch/issues/89} and titled as "Launching on arm64 with Fast-RTPS with fat archive from 2018-06-21 never quits". This bug affects a specific implementation of the communication middleware used in ROS 2.0 (DDS) and causes the middleware to hang after terminating the execution of a launch file (an abstraction used to simplify the launch of processes and tasks in ROS 2.0). This can potentially lead to several vulnerabilities such as leaking information or representing a point of access to potential attackers from whereto escalate into the control of critical systems within the robot.

If we were to assume the former, this bug causes a vulnerability that allows an attacker to take control of the robot. Based on this assumption, let's calculate the corresponding CVSSv3 scoring:

\begin{equation}
  \label{eq:eqd}
    \resizebox{.9\hsize}{!}{$
    CVSS:3.0/\underbrace{\underbrace{\underbrace{\underbrace{AV:N }_\text{0.85}/\underbrace{AC:L}_\text{0.77}/\underbrace{PR:N}_\text{0.85}/\underbrace{UI:N}_\text{0.85}}_\text{Exploitability = 8.22$\cdot$0.85$\cdot$0.77$\cdot$0.85$\cdot$0.85 = 3.887042775}
    /\underbrace{\underbrace{S:U}_\text{Unchanged}/\underbrace{\underbrace{C:N}_\text{0}/\underbrace{I:N}_\text{0}/\underbrace{A:N}_\text{0}}_\text{1-[(1-0)$\cdot$(1-0)$\cdot$(1-0)]}}_\text{Impact = 0}}_\text{$Base_{score}$  = roundup(min[(0), 10]) = roundup(0) = 0}}_\text{CVSSv3 = 0}
    $}
\end{equation}

Note that \emph{Availability}, \emph{Integrity} and \emph{Confidentiality} have all been marked to zero (0) since none of these aspects are compromised when evaluated with respect a vulnerability in the middleware. This results into an \emph{Impact} metric subgroup of 0 which according to the mathematical background of CVSSv3, results in a 0 total score.

As suggested by Carlage\cite{cvss3improvements}, an approach to generating a more useful CVSS score is to guess how third party code commonly calls the library, and score for the reasonable worst case. According to the author, this is easier for libraries like OpenSSL that are typically used for a defined purpose, in this case securing network communication, but it is less simple for other types of library.

For robotics, rating middleware vulnerabilities is critical as it's often the underlying layer used to realize whatever functionality to be implemented. We adopt Carlage's thesis and evaluate how third party libraries and middleware may be used and scored according to a reasonable worst case scenario. 

In the context of our previous example, applying the modifications introduced in RVSS for rating libraries:

\begin{equation}
  \label{eq:eqd2}
    \resizebox{.9\hsize}{!}{$
    RVSS:1.0/\underbrace{\underbrace{\underbrace{\underbrace{AV:AN}_\text{0.62}/\underbrace{AC:L}_\text{0.77}/\underbrace{PR:N}_\text{0.85}/\underbrace{UI:N}_\text{0.85}/\underbrace{Y:O}_\text{1.1}}_\text{Exploitability = 8.22$\cdot$0.62$\cdot$0.77$\cdot$0.85$\cdot$0.85$\cdot$1.1 = 3.118780203}
    /\underbrace{\underbrace{S:U}_\text{Unchanged}/\underbrace{\underbrace{C:N}_\text{0}/\underbrace{I:L}_\text{0}/\underbrace{A:N}_\text{0}/\underbrace{H:H}_\text{0.35}}_\text{1-[(1-0)$\cdot$(1-0)$\cdot$(1-0)] + 1.2$\cdot$0.35 = 0.42}}_\text{Impact = 6.42$\cdot$0.42 = 2.6964}}_\text{$Base_{score}$  = roundup(min[(3.118780203 + 2.6964), 10]) = roundup(5.815180203) = 5.9}}_\text{RVSS = 5.9}
    $}
\end{equation}

which delivers a final score of 5.9. Note the modification in \emph{Safety (H) = H} metric that affects the \emph{Impact} metric subgroup. The rationale of this modification follow from Carlage's suggestion to \emph{"score for the reasonable worst case"}\footnote{Some could argue that the \emph{Integrity (I)} metric should be re-evaluated resulting in a vector like \emph{RVSS:1.0/AV:AN/AC:L/PR:N/UI:N/Y:O/S:U/C:N/I:L/A:N/H:H} with a rating of 7.3}.


\section{Conclusions and future work}
\label{sec:conclusions}

We proposed herein to the community of roboticists, security researchers and penetration testers the foundations of the Robot Vulnerability Scoring System (RVSS)  version v1.0. Our research work draws some major conclusions: 

First, we highly value CVSS(v3) as an original and pioneering effort into vulnerability logging and registry, providing and adequate cataloguing system, which was an extremely valid for tool for IT environments for more than a decade. Unfortunately, CVSSv3 is not accurate enough to assess severity of vulnerabilities affecting an array of technologies that have direct or indirect physical implications on the environment, such as robotics. While this work was greatly inspired by the former, we have shown evidence that cyberphysical systems, but robotics in particular, deserve a separate consideration. 


Secondly, the present research work aims to capture the complexity of robot vulnerabilities in a Robot Vulnerability Scoring System (RVSS) version 1, presented herein. The main contributions of RVSS are a) the need of addressing \emph{Safety} and \emph{Environment} through a specific metrics, b) the modifications in the \emph{Attack Vector} to reflect robot specific scenarios, c) the addition of an \emph{Age} modifier to capture the relevance of aged, non-catalogued and non-fixed vulnerabilities and d) the proposal of changes and new interpretations to obtain non-zero scores for libraries and third-party software components, particular and increasingly critical in the robotics ecosystem. The specifications mentioned have been integrated and formulated\footnote{see Section \ref{appendix:rvssmath}.} as to allow retrospective compatibility and simplified comparison to CVSSv3. Accordingly, RVSS may be used almost effortless to review robot vulnerabilities reported within the CVSS scheme.

Thirdly, by the virtue of the research piece herein, we provide clearcut evidence of the relevance of considering particular robot deployment contexts when assessing robot security, and we believe that RVSS needs to converge towards a exhaustive characterization of the environment.
 
Finally, we conducted a experimental performance validation and assessment of RVSS, in comparison with CVSSv3 that validates our preliminary hypotheses. Importantly, Safety is identified as the differentiating factor between most conventional and robot related vulnerabilities; but context dependent data also adds great value to  severity assessment, as exemplified through a variety of use cases.

We foresee further collaborative efforts to improve and extend research on robot vulnerability evaluation. Accordingly, we provide reference RVSS material freely accessible and available. An open source implementation of the scoring system is available at \url{http://github.com/aliasrobotics/RVSS} licensed under GPLv3. We kindly invite security researchers, robotics researchers, pentesters and analysts to review, challenge and complement the present piece of research.  Lastly, we encourage the robot hacker community the use of RVSS v1.0 for robot vulnerability assessment.


\section*{Acknowledgements}

This research has been partially funded by the Basque Government, in particular, by the Business Development Basque Agency (SPRI) through the \emph{Ekintzaile} 2018 program. Special thanks to BIC Araba and the Basque Cybersecurity Center (BCSC) for their support. 



\bibliography{iclr2018_workshop}
\bibliographystyle{iclr2018_workshop}

\newpage
\appendix

\section{CVSS v3.0 Qualitative Severity Rating Scale}
\label{appendix:cvss3severity}

\ra{2}
\begin{table}[h!]
  \centering
  \caption{CVSS version 3 qualitative severity rating scale.}
  \label{table:cvss3severity}
  \begin{tabular}{|l c|}
  \textbf{Rating} & \textbf{CVSS score} \\ \hline
  \rowcolor{green!8} None & 0 \\
  \rowcolor{yellow!8} Low & 0.1 - 3.9\\
  \rowcolor{orange!8} Medium & 4.0 - 6.9\\
  \rowcolor{red!8} High & 7.0 - 8.9\\
  \rowcolor{red!20} Critical & 9.0 - 10.0\\
  \end{tabular}
\end{table}

\section{CVSS v3.0 specification}
\label{appendix:cvss3summary}

\ra{3}
\begin{longtable}{|l p{2cm} p{2cm} p{4cm} l|}
    \caption{Common Vulnerability Scoring System (CVSS) version 3.0 specification.}
    \label{table:cvss3}\\ 
    \toprule
\textbf{Metric Group} & \textbf{Subgroup} & \textbf{Metric Name } &  \textbf{Possible values} & \textbf{Mandatory} \\
    \midrule
    \endfirsthead 
    \toprule
\textbf{Metric Group} & \textbf{Subgroup} & \textbf{Metric Name} &  \textbf{Possible values} & \textbf{Mandatory} \\
    \midrule
    \endhead 
\rowcolor{black!5} Base & Exploitability & Attack Vector (\textbf{AV}) & \texttt{Network (N) = 0.85, Adjacent Network (A) = 0.62, Local (L) = 0.55, Physical (P) = 0.2} & Yes \\

 & & Attack Complexity (\textbf{AC}) & \texttt{Low (L) = 0.77}, \texttt{High (H) = 0.44} & Yes\\

\rowcolor{black!5} & & Privileges Required (\textbf{PR}) & \texttt{None (N) = 0.85, Low (L) = 0.62 (0.68 if scope / modified scope is changed), High (H) = 0.27 (0.50 if scope / modified scope is changed)} & Yes\\

 & & User Interaction (\textbf{UI}) & \texttt{None (N) = 0.85, Required (R) = 0.62} & Yes \\

\rowcolor{black!5} & & Scope (\textbf{S}) & \texttt{Unchanged (U), Changed (C)} & Yes\\

 & Impact & Confidentiality (\textbf{C}) &  \texttt{High (H) = 0.56, Low (L) = 0.22, None (N) = 0} & Yes\\

\rowcolor{black!5} & & Integrity (\textbf{I}) & \texttt{High (H) = 0.56, Low (L) = 0.22, None (N) = 0} & Yes\\

 & & Availability (\textbf{A}) & High (H) = 0.56, Low (L) = 0.22, None (N) = 0 & Yes\\

\rowcolor{black!5} Temporal & & Exploit Code Maturity (\textbf{E}) & \texttt{Not Defined (X) = 1, High (H) = 1, Functional (F) = 0.97, Proof of Concept (P) =  0.94, Unproven (U) = 0.91} & No\\

 & & Remediation Level (\textbf{RL}) & \texttt{Not Defined (X) = 1, Unavailable (U) = 1, Workaround (W) = 0.97, Temporary Fix (T) = 0.96, Official Fix (O) = 0.95} & No\\

\rowcolor{black!5} &  & Report Confidence (\textbf{RC}) & \texttt{Not Defined (X) = 1, Confirmed (C) = 1, Reasonable (R) 0.96, Unknown (U) = 0.92} & No\\

 Environmental &  & Confidentiality Requirement (\textbf{CR}) & \texttt{Not Defined (X) = 1, Low (L) = 0.5, Medium (M) = 1, High (H) = 1.5} & No\\

\rowcolor{black!5}  &  & Integrity Requirement (\textbf{IR}) & \texttt{Not Defined (X) = 1, Low (L) = 0.5, Medium (M) = 1, High (H) = 1.5} &  No\\

  &  & Availability Requirement (\textbf{AR}) & \texttt{Not Defined (X) = 1, Low (L) = 0.5, Medium (M) = 1, High (H) = 1.5} & No\\

 \rowcolor{black!5}  & Exploitability & Modified Attack Vector (\textbf{MAV}) & \texttt{Not Defined (X) = 1,Network (N) = 0.85,Adjacent Network (A) = 0.62, Local (L) = 0.55, Physical (P) = 0.2} & No\\

  & & Modified Attack Complexity (\textbf{MAC}) & \texttt{Not Defined (X) = 1,Low (L) = 0.77, High (H) = 0.44} & No\\

\rowcolor{black!5}  & & Modified Privileges Required (\textbf{MPR}) & \texttt{Not Defined (X) = 1,None (N) = 0.85, Low (L) = 0.62 (0.68 if scope / modified scope is changed), High (H) = 0.27 (0.50 if scope / modified scope is changed)} & No\\

 & & Modified User Interaction (\textbf{MUI}) & \texttt{Not Defined (X) = 1,None (N) = 0.85, Required (R) = 0.62} & No\\

\rowcolor{black!5} & Impact & Modified Scope (\textbf{MS}) & \texttt{Not Defined (X) = 1,Unchanged (U), Changed (C)} & No\\

 & & Modified Confidentiality (\textbf{MC}) &  \texttt{Not Defined (X) = 1,High (H) = 0.56, Low (L) = 0.22, None (N) = 0} & No\\

\rowcolor{black!5}  & & Modified Integrity (\textbf{MI}) & \texttt{Not Defined (X) = 1,High (H) = 0.56, Low (L) = 0.22, None (N) = 0} & No\\

 & & Modified Availability (\textbf{MA}) & \texttt{Not Defined (X) = 1,High (H) = 0.56, Low (L) = 0.22, None (N) = 0} & No\\
    \bottomrule
\end{longtable}

\newpage
\section{RVSS v1.0 specification}
\label{appendix:rvss1summary}

\ra{2}
\begin{longtable}{|l l p{2cm} p{3cm} p{4cm} |}
    \caption{Robot Vulnerability Scoring System (CVSS) version 1.0 specification.}
    \label{table:rvss}\\ 
    \toprule
\textbf{Group} & \textbf{Subgroup} & \textbf{Metric Name } & \textbf{Description} &  \textbf{Possible values} \\
    \midrule
    \endfirsthead 
    \toprule
\textbf{Group} & \textbf{Subgroup} & \textbf{Metric Name } & \textbf{Description} &  \textbf{Possible values} \\
    \midrule
    \endhead 


\rowcolor{bluealias!25} Base & Exploitability & Attack Vector (\textbf{AV}) & \textit{context by which vulnerability exploitation is possible (larger the more remote)} & \texttt{Remote Network (RN) = 0.85, Adjacent Network (AN) = 0.62, Internal Network (IN) = 0.4, Local (L) = 0.55, Physical Public (PP) = 0.62, Physical Restricted (PR) = 0.4, Physical Isolated (PI) = 0.2} and combinations as described above.\\

\rowcolor{bluealias!25} & & Attack Complexity (\textbf{AC}) & \textit{conditions beyond the attacker’s control that must exist in order to exploit the vulnerability} & \texttt{Low (L) = 0.77}, \texttt{High (H) = 0.44}\\

\rowcolor{bluealias!25} & & Privileges Required (\textbf{PR}) & \textit{level of privileges an attacker must possess before successfully exploiting the vulnerability} & \texttt{None (N) = 0.85, Low (L) = 0.62, High (H) = 0.27}\\

\rowcolor{bluealias!25} & & User Interaction (\textbf{UI}) & \textit{requirement for a user, other than the attacker, to participate in the successful compromise of the vulnerable component} & \texttt{None (N) = 0.85, Required (R) = 0.62} \\

\rowcolor{bluealias!25} & & Age (\textbf{Y}) & \textit{timespan since vulnerability was first reported (in years)} & \texttt{Zero Day (Z) = 1.0, 1 year or less (O) = 1.1, Less than 3 years (T) = 1.3, More than 3 years (M) = 1.5, Unknown (U) = 1.0} \\

\rowcolor{bluealias!25} & & Scope (\textbf{S}) & \textit{ability for a vulnerability in one software component to impact resources beyond its means, or privileges} & \texttt{Unchanged (U), Changed (C)} \\

\rowcolor{bluealias!35} & Impact & Confidentiality (\textbf{C}) & \textit{measures the impact to the confidentiality\footnote{confidentiality refers to limiting information access and disclosure to only authorized users} of the information resources managed by the robot, the robot module or the component} & \texttt{High (H) = 0.56, Low (L) = 0.22, None (N) = 0} \\

\rowcolor{bluealias!35} & & Integrity (\textbf{I}) & \textit{measures the impact to integrity\footnote{integrity refers to the trustworthiness and veracity of information} of a successfully exploited vulnerability} & \texttt{High (H) = 0.56, Low (L) = 0.22, None (N) = 0} \\

\rowcolor{bluealias!35} & & Availability (\textbf{A}) & \textit{measures the impact to the availability\footnote{availability refers to the accessibility of information resources} of the impacted component resulting from a successfully exploited vulnerability} & \texttt{High (H) = 0.56, Low (L) = 0.22, None (N) = 0} \\

\rowcolor{bluealias!35} & & Safety (\textbf{H}) & \textit{measures potential physical hazards on humans or on the environment} & \texttt{Unknown (U) = 0.0, None (N) = 0.0, Environmental (E) = 0.15, Human (HU) = 0.35} \\


\rowcolor{bluealias!15} Temporal & & Exploit Code Maturity (\textbf{E}) &  \textit{measures the likelihood of the vulnerability being attacked based on the current state of exploit techniques} & \texttt{Not Defined (X) = 1, High (H) = 1, Functional (F) = 0.97, Proof of Concept (P) =  0.94, Unproven (U) = 0.91} \\

\rowcolor{bluealias!15} & & Remediation Level (\textbf{RL}) & \textit{measures the remediation state of a vulnerability, the less official and permanent a fix, the higher the vulnerability score} & \texttt{Not Defined (X) = 1, Unavailable (U) = 1, Workaround (W) = 0.97, Temporary Fix (T) = 0.96, Official Fix (O) = 0.95} \\

\rowcolor{bluealias!15} &  & Report Confidence (\textbf{RC}) & \textit{measures the degree of confidence in the existence of the vulnerability and the credibility of the known technical details} & \texttt{Not Defined (X) = 1, Confirmed (C) = 1, Reasonable (R) 0.96, Unknown (U) = 0.92} \\

\rowcolor{bluealias!5} Environmental &  & Confidentiality Requirement (\textbf{CR}) & \textit{enables the analyst to customize score depending on the importance of confidentiality}  & \texttt{Not Defined (X) = 1, Low (L) = 0.5, Medium (M) = 1, High (H) = 1.5} \\

 \rowcolor{bluealias!5}  &  & Integrity Requirement (\textbf{IR}) & \textit{enables the analyst to customize score depending on the importance of integrity} & \texttt{Not Defined (X) = 1, Low (L) = 0.5, Medium (M) = 1, High (H) = 1.5} \\

 \rowcolor{bluealias!5} &  & Availability Requirement (\textbf{AR}) & \textit{enables the analyst to customize score depending on the importance of availability} & \texttt{Not Defined (X) = 1, Low (L) = 0.5, Medium (M) = 1, High (H) = 1.5} \\

\rowcolor{bluealias!5} &  & Safety Requirement (\textbf{HR}) & \textit{enables the analyst to customize score depending on the importance of safety} & \texttt{Not Defined (X) = 1, Low (L) = 0.5, Medium (M) = 1, High (H) = 1.5} \\

 \rowcolor{black!5}  & Exploitability & Modified Attack Vector (\textbf{MAV}) & \textit{enable the analyst to adjust the base metrics according to modifications that exist within the analyst’s environment} & \texttt{Not Defined (X) = 1,Network (N) = 0.85,Adjacent Network (A) = 0.62, Local (L) = 0.55, Physical (P) = 0.2} \\

 \rowcolor{black!5} & & Modified Attack Complexity (\textbf{MAC}) & \textit{enable the analyst to adjust the base metrics according to modifications that exist within the analyst’s environment} &  \texttt{Not Defined (X) = 1,Low (L) = 0.77, High (H) = 0.44} \\

\rowcolor{black!5}  & & Modified Privileges Required (\textbf{MPR}) & \textit{enable the analyst to adjust the base metrics according to modifications that exist within the analyst’s environment} & \texttt{Not Defined (X) = 1, None (N) = 0.85, Low (L) = 0.68, High (H) = 0.50} \\

\rowcolor{black!5} & & Modified User Interaction (\textbf{MUI}) & \textit{enable the analyst to adjust the base metrics according to modifications that exist within the analyst’s environment} & \texttt{Not Defined (X) = 1, None (N) = 0.85, Required (R) = 0.62} \\

\rowcolor{black!5} & & Modified Age (\textbf{MY}) & \textit{enable the analyst to adjust the base metrics according to modifications that exist within the analyst’s environment} & \texttt{Not Defined (X) = 1, Zero Day (Z) = 1.0, 1 year or less (O) = 1.1, Less than 3 years (T) = 1.3, More than 3 years (M) = 1.5, Unknown (U) = 1.0} \\

\rowcolor{black!8} & Impact & Modified Scope (\textbf{MS}) & \textit{enable the analyst to adjust the base metrics according to modifications that exist within the analyst’s environment} & \texttt{Not Defined (X) = 1.0, Unchanged (U), Changed (C)} \\

\rowcolor{black!8} & & Modified Confidentiality (\textbf{MC}) & \textit{enable the analyst to adjust the base metrics according to modifications that exist within the analyst’s environment} & \texttt{Not Defined (X) = 1.0, High (H) = 0.56, Low (L) = 0.22, None (N) = 0} \\

\rowcolor{black!8}  & & Modified Integrity (\textbf{MI}) & \textit{enable the analyst to adjust the base metrics according to modifications that exist within the analyst’s environment} & \texttt{Not Defined (X) = 1.0, High (H) = 0.56, Low (L) = 0.22, None (N) = 0} \\

\rowcolor{black!8} & & Modified Availability (\textbf{MA}) & \textit{enable the analyst to adjust the base metrics according to modifications that exist within the analyst’s environment} & \texttt{Not Defined (X) = 1.0, High (H) = 0.56, Low (L) = 0.22, None (N) = 0} \\

\rowcolor{black!8} & & Modified Safety (\textbf{MH}) & \textit{enable the analyst to adjust the base metrics according to modifications that exist within the analyst’s environment, in particular, regarding safety aspects} & \texttt{Not Defined (X) = 1.0, Unknown (U) = 0.0, None (N) = 0.0, Environmental (E) = 0.56, Human (HU) = 0.8} \\

    \bottomrule
\end{longtable}

\newpage
\section{CVSS v3.0 mathematical foundation}
\label{appendix:cvss3math}

\subsection{Base}
\label{appendix:cvss3:base}

The $Exploitability$ metric is calculated in Equation \ref{eq:exploitability}:

\begin{equation}
\begin{split}
    Explotaibility = 8.22 \cdot AV \cdot AC
    \cdot PR \cdot UI
\end{split}
\label{eq:exploitability}
\end{equation}


The $Impact$ metric is determined through Equations \ref{eq:impact} and \ref{eq:impact2}:
\begin{equation}
    ISC_{Base} = 1 - [(1 - C) \cdot (1 - I ) \cdot (1 - A)]
\label{eq:impact}
\end{equation}

\begin{equation}
\resizebox{.8\hsize}{!}{$
Impact =
\begin{cases}
  6.42 \cdot ISC_{Base}, & \text{if}\ Scope \text{ Unchanged}\\
  7.52 \cdot [ISC_{Base} - 0.029] - 3.25 \cdot [ISC_{Base} - 0.02]^{15}, & \text{if}\ Scope \text{ Changed}\\
\end{cases}
$}
\label{eq:impact2}
\end{equation}

Then, altogether, the $Base_{score}$ metric can be calculated through Equation \ref{eq:base}:
\begin{equation}
\resizebox{.8\hsize}{!}{$
Base_{score} =
\begin{cases}
  0, & \text{if}\ Impact \text{ subscore}\ <=0\\
  roundup(min[(Impact + Exploitability), 10]), & \text{if}\ Scope \text{ Unchanged}\\
  roundup(min[1.08 \cdot ( Impact+Exploitability) , 10]), & \text{if}\ Scope \text{ Changed}\\
\end{cases}
$}
\label{eq:base}
\end{equation}

where $roundup()$ is defined as the smallest number, specified to one decimal place, that is equal to or higher than its input. For example, $roundup(4.02)= 4.1$ and $roundup(4.00)=4.0$.

\subsection{Temporal}
\label{appendix:cvss3:temporal}

The $Temporal$ score is defined in Equation \ref{eq:temporal}:

\begin{equation}
Temporal = roundup(Base_{score} \cdot  E
\cdot RL \cdot RC)
\label{eq:temporal}
\end{equation}

\subsection{Environmental}
\label{appendix:cvss3:environmental}

The modified exploitability $M.Exploitability$  score is:

\begin{equation}
    M.Explotaibility = 8.22 \cdot MAV \cdot MAC
    \cdot MPR \cdot MUI
\end{equation}

and the modified impact $M.Impact$ score is defined as:

\begin{equation}
\resizebox{.8\hsize}{!}{$
M.Impact =
\begin{cases}
  6.42 \cdot ISC_{Modified}, & \text{if Modified}\ Scope \text{ Unchanged}\\
  7.52 \cdot [ISC_{Modified} - 0.029]  & \text{if Modified}\ Scope \text{ Changed}\\
  - 3.25 \cdot [ISC_{Modified} - 0.02]^{15} & \\
\end{cases}
$}
\label{eq:mimpact}
\end{equation}

where

\begin{equation}
\resizebox{.8\hsize}{!}{$
ISC_{Modified}= min\bigg(\Big[1 - (1 - MC \cdot CR)
\cdot (1 - MI \cdot IR)
\cdot (1 - MA \cdot AR)\Big],0.915 \bigg)
$}
\end{equation}

which ends up into:

\begin{equation}
\resizebox{.8\hsize}{!}{$
Environmental =
\begin{cases}
  0, & \text{if modified}\ Impact <= 0 \\
  & \\
roundup\bigg(roundup\Big(min[ (M.Impact + M.Exploitability) ,10]\Big) \cdot E \cdot  RL \cdot RC\bigg) & \text{if Modified}\ Scope \text{ is Unchanged} \\
  & \\
roundup\bigg(roundup\Big(min[1.08 \cdot (M.Impact + M.Exploitability) ,10]\Big) \cdot E \cdot RL   \cdot RC\bigg) & \text{if Modified}\ Scope \text{ is Changed} \\
\end{cases}
$}
\label{eq:base}
\end{equation}

\newpage
\section{RVSS v1.0 mathematical foundation}
\label{appendix:rvssmath}

\subsection{Base}
\label{appendix:cvss3:base}

The $Exploitability$ metric is calculated in Equation \ref{eq:exploitability}:

\begin{equation}
\begin{split}
    Explotaibility = 8.22 \cdot AV \cdot AC
    \cdot PR \cdot UI \cdot Y
\end{split}
\label{eq:exploitability}
\end{equation}


The $Impact$ metric is determined through Equations \ref{eq:impact} and \ref{eq:impact2}:


\begin{equation}
    ISC_{Base} = 1 - [(1 - C) \cdot (1 - I ) \cdot (1 - A)] + 1.2 \cdot H
\label{eq:impact}
\end{equation}

\begin{equation}
\resizebox{.8\hsize}{!}{$
Impact =
\begin{cases}
  6.42 \cdot ISC_{Base}, & \text{if}\ Scope \text{ Unchanged}\\
  7.52 \cdot [ISC_{Base} - 0.029] - 3.25 \cdot [ISC_{Base} - 0.02]^{15}, & \text{if}\ Scope \text{ Changed and } ISC_{Base} < 1.0\\
  7.52 \cdot [ISC_{Base} - 0.029] - 3.25 \cdot [1.0 - 0.04]^{15}, & \text{if}\ Scope \text{ Changed and } ISC_{Base} > 1.0\\
  
\end{cases}
$}
\label{eq:impact2}
\end{equation}

Then, altogether, the $Base_{score}$ metric can be calculated through Equation \ref{eq:base}:
\begin{equation}
\resizebox{.8\hsize}{!}{$
Base_{score} =
\begin{cases}
  0, & \text{if}\ Impact \text{ subscore}\ <=0\\
  roundup(min[(Impact + Exploitability), 10]), & \text{if}\ Scope \text{ Unchanged}\\
  roundup(min[1.08 \cdot ( Impact+Exploitability) , 10]), & \text{if}\ Scope \text{ Changed}\\
\end{cases}
$}
\label{eq:base}
\end{equation}

where $roundup()$ is defined as the smallest number, specified to one decimal place, that is equal to or higher than its input. For example, $roundup(4.02)= 4.1$ and $roundup(4.00)=4.0$.

\subsection{Temporal}
\label{appendix:cvss3:temporal}

The $Temporal$ score is defined in Equation \ref{eq:temporal}:

\begin{equation}
Temporal = roundup(Base_{score} \cdot  E
\cdot RL \cdot RC)
\label{eq:temporal}
\end{equation}

\subsection{Environmental}
\label{appendix:cvss3:environmental}

The modified exploitability $M.Exploitability$  score is:

\begin{equation}
    M.Explotaibility = 8.22 \cdot MAV \cdot MAC
    \cdot MPR \cdot MUI
\end{equation}

and the modified impact $M.Impact$ score is defined as:

\begin{equation}
\resizebox{.8\hsize}{!}{$
M.Impact =
\begin{cases}
  6.42 \cdot ISC_{Modified}, & \text{if}\ Scope \text{ Unchanged}\\
  7.52 \cdot [ISC_{Modified} - 0.029] - 3.25 \cdot [ISC_{Modified} - 0.02]^{15}, & \text{if}\ Scope \text{ Changed and } ISC_{Base} < 1.0\\
  7.52 \cdot [ISC_{Modified} - 0.029] - 3.25 \cdot [1.0 - 0.04]^{15}, & \text{if}\ Scope \text{ Changed and } ISC_{Base} > 1.0\\

\end{cases}
$}
\label{eq:mimpact}
\end{equation}

where


\begin{equation}
\resizebox{.8\hsize}{!}{$
ISC_{Modified}= min\bigg(\Big[1 - (1 - MC \cdot CR)
\cdot (1 - MI \cdot IR)
\cdot (1 - MA \cdot AR) \cdot (1 - MH \cdot HR)\Big],0.915 \bigg)
+ 1.2 \cdot MH \cdot HR
$}
\end{equation}

which ends up into:

\begin{equation}
\resizebox{.8\hsize}{!}{$
Environmental =
\begin{cases}
  0, & \text{if modified}\ Impact <= 0 \\
  & \\
roundup\bigg(roundup\Big(min[ (M.Impact + M.Exploitability) ,10]\Big) \cdot E \cdot  RL \cdot RC\bigg) & \text{if Modified}\ Scope \text{ is Unchanged} \\
  & \\
roundup\bigg(roundup\Big(min[1.08 \cdot (M.Impact + M.Exploitability) ,10]\Big) \cdot E \cdot RL   \cdot RC\bigg) & \text{if Modified}\ Scope \text{ is Changed} \\
\end{cases}
$}
\label{eq:base}
\end{equation}

\end{document}